\documentclass{article} %
\usepackage{nips14submit_e,times}
\usepackage{hyperref}
\usepackage{url}
\usepackage{booktabs}
\usepackage{subfig}
\usepackage{graphicx} %
\usepackage{microtype}

\title{Document Embedding with Paragraph Vectors}

\author{
Andrew M. Dai \\
Google\\
\texttt{adai@google.com} \\
\And
Christopher Olah \\
Google\\
\texttt{colah@google.com} \\
\And
Quoc V. Le \\
Google\\
\texttt{qvl@google.com} \\
}

\nipsfinalcopy %

\begin{document}

\maketitle

\begin{abstract}
Paragraph Vectors has been recently proposed as an unsupervised method
for learning distributed representations for pieces of texts. In their
work, the authors showed that the method can learn an embedding of
movie review texts which can be leveraged for sentiment analysis. That
proof of concept, while encouraging, was rather narrow.  Here we
consider tasks other than sentiment analysis, provide a more thorough
comparison of Paragraph Vectors to other document modelling algorithms
such as Latent Dirichlet Allocation, and evaluate performance of the
method as we vary the dimensionality of the learned representation. We
benchmarked the models on two document similarity data sets, one from
Wikipedia, one from arXiv. We observe that the Paragraph Vector method
performs significantly better than other methods, and propose a
simple improvement to enhance embedding quality.  Somewhat
surprisingly, we also show that much like word embeddings, vector
operations on Paragraph Vectors can perform useful semantic
results.

\end{abstract}

\section{Introduction}
Central to many language understanding problems is the question of
knowledge representation: How to capture the essential meaning of a
document in a machine-understandable format (or
``representation''). Despite much work going on in this area, the most
established format is perhaps the bag of words (or bag of n-gram)
representations~\cite{harris54}. Latent Dirichlet Allocation
(LDA)~\cite{blei03} is another widely adopted representation.

A recent paradigm in machine intelligence is to use a distributed
representation for words~\cite{mikolov} and documents~\cite{le14}. The
interesting part is that even though these representations are less
human-interpretable than previous representations, they seem to work well
in practice. In particular, Le and Mikolov~\cite{le14} show that
their method, Paragraph Vectors, capture many document semantics
in dense vectors and that they can be used in classifying movie reviews or
retrieving web pages.

Despite their success, little is known about how well the model works
compared to Bag-of-Words or LDA for other unsupervised applications and how
sensitive the model is when we change the hyperparameters. 

In this paper, we make an attempt to compare Paragraph Vectors with
other baselines in two tasks that have significant practical
implications. First, we benchmark Paragraph Vectors on the task of
Wikipedia browsing: given a Wikipedia article, what are the nearest
articles that the audience should browse next. We also test Paragraph
Vectors on the task of finding related articles on arXiv. In both of
these tasks, we find that Paragraph Vectors allow for finding
documents of interest via simple and intuitive vector operations. For example, we can
find the Japanese equivalence of ``Lady Gaga.''. 

The goal of the paper is beyond benchmarking: The positive results on
Wikipedia and arXiv datasets confirm that having good representations
for texts can be powerful when it comes to language understanding. The
success in these tasks shows that it is possible to use Paragraph
Vectors for local and non-local browsing of large corpora.

We also show a simple yet effective trick to improve Paragraph
Vector. In particular, we observe that by jointly training word embeddings, as in the skip gram model,
the quality of the paragraph vectors is improved.

\section{Model}
The Paragraph Vector model is first proposed in~\cite{le14}. The model
inserts a memory vector to the standard language model which aims at
capturing the topics of the document. The authors named this model
``Distributed Memory'':

\begin{figure}[h!]
\centering
\includegraphics[width=0.5\textwidth]{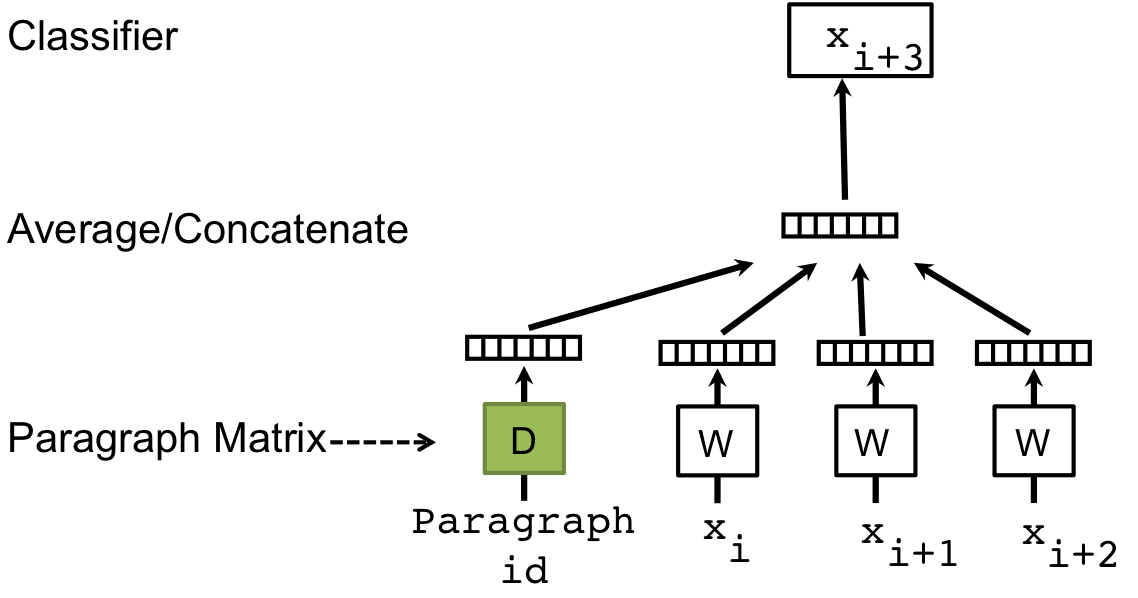}
\caption{The distributed memory model of Paragraph Vector for an input sentence.}
\end{figure}

As suggested by the figure above, the paragraph vector is concatenated
or averaged with local context word vectors to predict the next
word. The prediction task changes the word vectors and the paragraph
vector.

The paragraph vector can be further simplified when we use no local
context in the prediction task. We can arrive at the following
``Distributed Bag of Words'' model:

\begin{figure}[h!]
\centering
\includegraphics[width=0.5\textwidth]{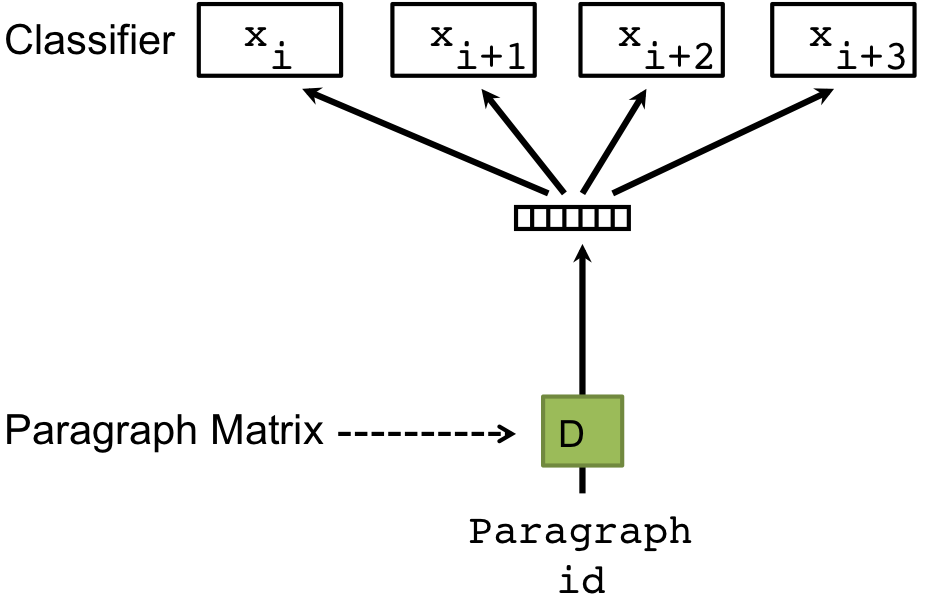}
\caption{The distributed bag of words model of Paragraph Vector.}
\end{figure}

At inference time, the parameters of the classifier and the word
vectors are not needed and backpropagation is used to tune the paragraph
vectors.

As the distributed bag of words model is more efficient, the
experiments in this paper focuses on this implementation of the
Paragraph Vector. In the following sections, we will explore the use
of Paragraph Vectors in different applications in document
understanding.

\section{Experiments}
We conducted experiments with two different publicly available
corpora: a corpus from the repository of electronic preprints (arXiv),
and a corpus from the online encyclopaedia (Wikipedia).

In each case, all words were lower-cased before the datasets were
used. We also jointly trained word embeddings with the paragraph
vectors since preliminary experiments showed that this can improve the
quality of the paragraph vectors. Preliminary results also showed that
training with both unigrams and bigrams does not improve the quality of the
final vectors. We present a range of qualitative and quantitative
results. We give some examples of nearest neighbours to some Wikipedia
articles and arXiv papers as well as a visualisation of the space of
Wikipedia articles. We also show some examples of nearest neighbours
after performing vector operations.

For the quantitative evaluation, we attempt to measure how well
paragraph vectors represent semantic similarity of related
articles. We do this by constructing (both automatically and by hand)
triplets, where each triplet consists of a pair of items that are
close to each other and one item that is unrelated.

For the publicly available corpora we trained paragraph vectors over
at least 10 epochs of the data and use a hierarchical softmax constructed as a
Huffman tree as the classifier. We use cosine similarity as the
metric. We also applied LDA with Gibbs sampling and 500 iterations
with varying numbers of topics. For LDA, we set $\alpha$ to 0.1 and
used values of $\beta$ between 0.01 and 0.000001. We used the posterior
topic proportions for each paper with Hellinger distance to compute
the similarity between pairs of documents. For completeness, we also
include the results of averaging the word embeddings for each word in
a paper and using that as the paragraph vector. Finally, we consider the
classical bag of words model where each word is represented as a
one-hot vector weighted by TF-IDF and the document is represented by that vector, with comparisons done using cosine similarity.

\subsection{Performance of Paragraph Vectors on Wikipedia entries}
\begin{figure}[h!]
\centering
\includegraphics[width=.75\linewidth]{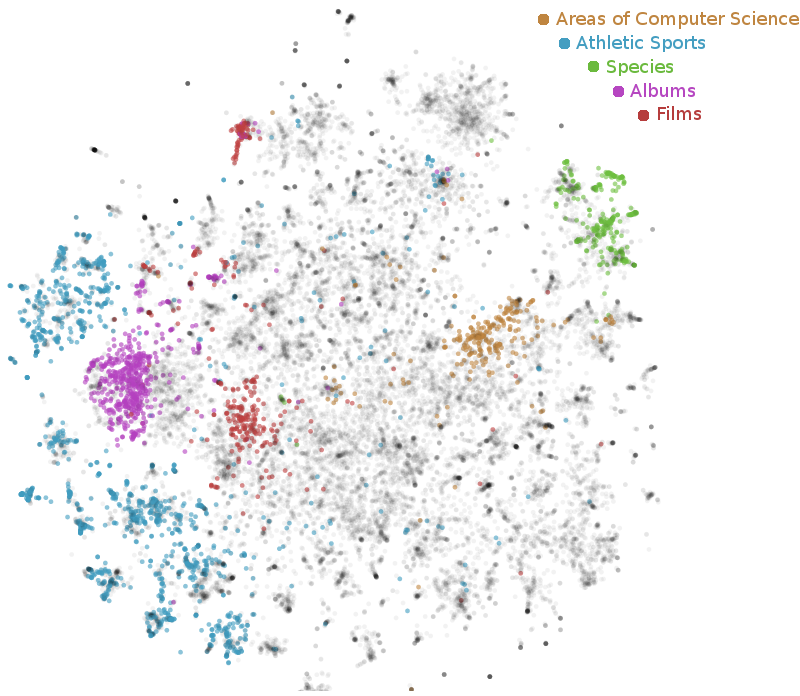}
\caption{Visualization of Wikipedia paragraph vectors using t-SNE. }
\label{wiki:vis}
\end{figure}

We extracted the main body text of 4,490,000 Wikipedia articles from
the English site. We removed all links and applied a frequency cutoff
to obtain a vocabulary of 915,715 words. We trained paragraph
vectors on these Wikipedia articles and visualized them in
Figure~\ref{wiki:vis} using t-SNE~\cite{maaten08}. The visualization
confirms that articles having the same category are grouped
together. There is a wide range of sport descriptions on wikipedia,
which explains why the sports are less concentrated.

We also qualitatively look at the nearest neighbours of Wikipedia
articles and compare Paragraph Vectors and LDA. For example, the
nearest neighbours for the Wikipedia article ``Machine learning'' are
shown in Table~\ref{tab:nn-ml}. We find that overall Paragraph Vectors
have better nearest neighbours than LDA.
\begin{table}[h!]
\caption{Nearest neighbours to ``Machine learning.'' Bold face texts
  are articles we found unrelated to ``Machine learning.'' We use
  Hellinger distance for LDA and cosine distance for Paragraph Vectors
  as they work the best for each model.}
\label{tab:nn-ml}
\begin{center}
\begin{tabular}{ll}
  \toprule
\multicolumn{1}{l}{\bf LDA } &\multicolumn{1}{l}{\bf Paragraph Vectors}
\\ \midrule
Artificial neural network & Artificial neural network	\\
Predictive analytics & Types of artificial neural networks \\
Structured prediction & Unsupervised learning \\
{\bf Mathematical geophysics} & Feature learning\\
Supervised learning & Predictive analytics \\
Constrained conditional model & Pattern recognition \\
Sensitivity analysis & Statistical classification\\
{\bf SXML} & Structured prediction\\
Feature scaling & Training set\\
Boosting (machine learning) & Meta learning (computer science)\\
Prior probability & Kernel method\\
Curse of dimensionality & Supervised learning \\
{\bf Scientific evidence} & Generalization error \\
Online machine learning & Overfitting \\
N-gram & Multi-task learning \\
Cluster analysis & Generative model \\
Dimensionality reduction & Computational learning theory\\
{\bf Functional decomposition} & Inductive bias\\
Bayesian network & Semi-supervised learning\\
\bottomrule
\end{tabular}
\end{center}
\end{table}

We can perform vector operations on paragraph vectors for local and
non-local browsing of Wikipedia. In Table~\ref{wiki-nn} and
Table~\ref{wiki-vec}, we show results of two experiments. The first
experiment is to find related articles to ``Lady Gaga.''  The second
experiment is to find the Japanese equivalence of ``Lady Gaga.'' This
can be achieved by vector operations: $pv$(``Lady Gaga'') -
$wv$(``American'') + $wv$(``Japanese'') where $pv$ is paragraph vectors and
$wv$ is word vectors. Both sets of results show that Paragraph Vectors
can achieve the same kind of analogies like Word
Vectors~\cite{mikolov}.

\begin{table}[h!]
\begin{center}
  \caption{Wikipedia nearest neighbours}
 \subfloat[Wikipedia nearest neighbours to ``Lady Gaga'' using Paragraph
  Vectors. All articles are relevant.] {
\label{wiki-nn}
\begin{tabular}{lc}
  \toprule
\multicolumn{1}{c}{\bf Article} &\multicolumn{1}{c}{\bf Cosine}\\
\multicolumn{1}{c}{\bf} &\multicolumn{1}{c}{\bf Similarity}
\\ \midrule
Christina Aguilera &	0.674 \\
Beyonce	& 0.645 \\
Madonna (entertainer)	& 0.643 \\
Artpop	 & 0.640 \\
Britney Spears	& 0.640 \\
Cyndi Lauper	& 0.632 \\
Rihanna	& 0.631 \\
Pink (singer)	& 0.628 \\
Born This Way	& 0.627 \\
The Monster Ball Tour	& 0.620  \\
\bottomrule
\end{tabular}
}
 \quad
\subfloat[Wikipedia nearest neighbours to ``Lady Gaga'' - ``American''
  + ``Japanese'' using Paragraph Vectors. Note that Ayumi Hamasaki is
  one of the most famous singers, and one of the best selling artists
  in Japan. She also has an album called ``Poker Face'' in 1998. ] {
\label{wiki-vec}
\begin{tabular}{lc}
  \toprule
\multicolumn{1}{c}{\bf Article} &\multicolumn{1}{c}{\bf Cosine}\\
\multicolumn{1}{c}{\bf } &\multicolumn{1}{c}{\bf Similarity}
\\ \midrule
Ayumi Hamasaki	& 0.539 \\
Shoko Nakagawa	& 0.531 \\
Izumi Sakai	& 0.512 \\
Urbangarde	& 0.505 \\
Ringo Sheena	& 0.503 \\
Toshiaki Kasuga	& 0.492 \\
Chihiro Onitsuka	& 0.487 \\
Namie Amuro	& 0.485 \\
Yakuza (video game)	& 0.485 \\
Nozomi Sasaki (model)	& 0.485 \\
\bottomrule
\end{tabular}
}
\end{center}
\end{table}

To quantitatively compare these methods, we constructed two datasets
for triplet evaluation. The first consists of 172 triplets of articles
we knew were related because of our domain knowledge. Some examples
are: ``Deep learning'' is closer to ``Machine learning'' than
``Computer network'' or ``Google'' is closer to ``Facebook'' than
``Walmart'' etc. Some examples are hard and probably require some deep
understanding of the content such as ``San Diego'' is closer to ``Los
Angeles'' than ``San Jose.''

The second dataset consists of 19,876 triplets in which two articles
are closer because they are listed in the same category by Wikipedia,
and the last article is not in the same category, but may be in a sibling category. For example, the articles for 
``Barack Obama'' are closer to ``Britney Spears'' than ``China.'' These
triplets are generated randomly.\footnote{The datasets are available at \url{http://cs.stanford.edu/~quocle/triplets-data.tar.gz}}

We will benchmark document embedding methods, such as LDA, bag of
words, Paragraph Vector, to see how well these models capture the
semantic of the documents. The results are reported in
Table~\ref{wikipedia-hand-table} and
Table~\ref{wikipedia-generated-table}.  For each of the methods, we
also vary the number of embedding dimensions.

\begin{figure}[h]
\centering
\includegraphics[width=.7\linewidth]{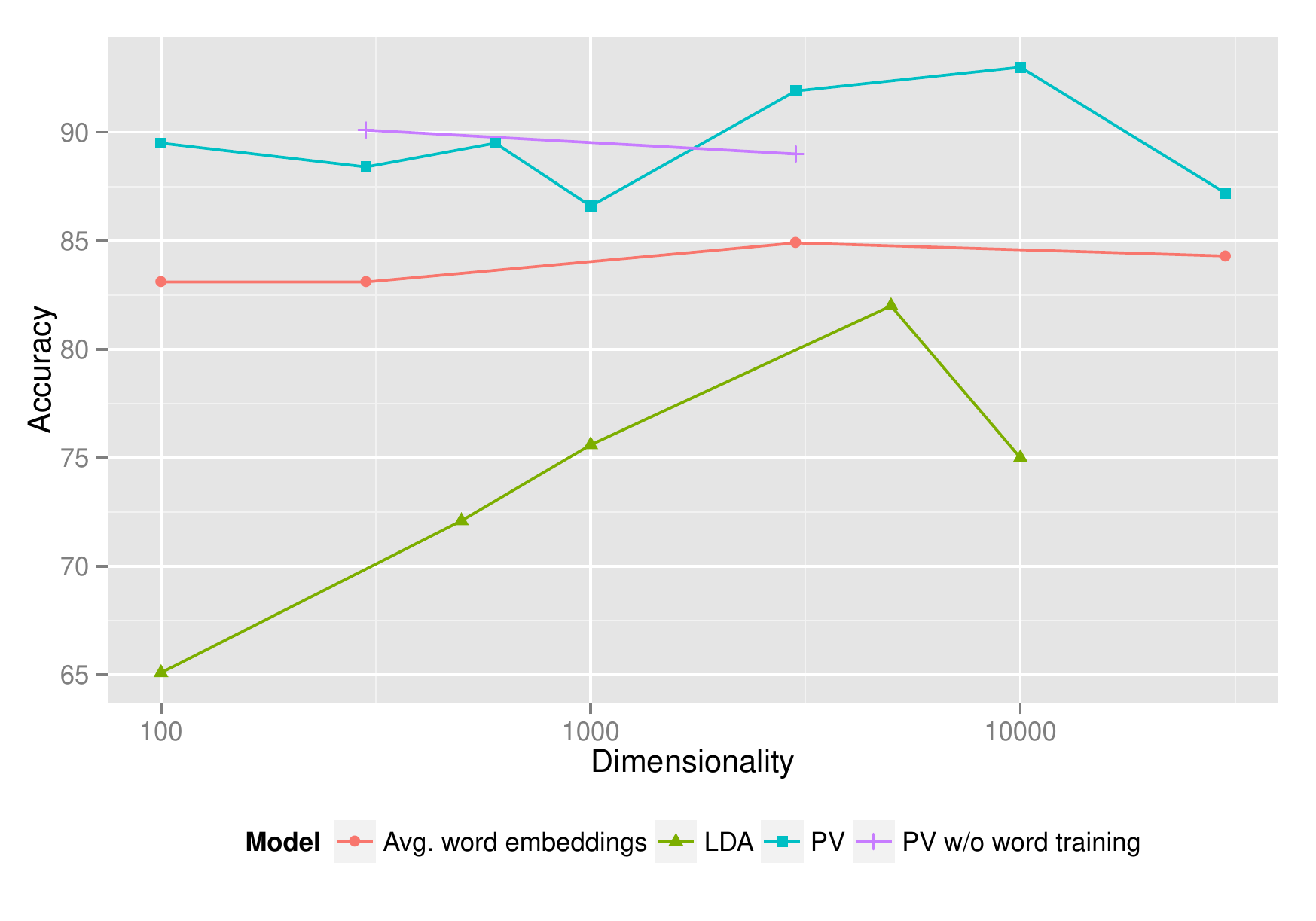}
\caption{Results of experiments on the hand-built Wikipedia triplet dataset. }
\label{wiki-hand-graph}
\end{figure}

\begin{table}[h!]
\caption{Performances of different methods on hand-built triplets of Wikipedia
  articles on the best performing dimensionality.}
\label{wikipedia-hand-table}
\begin{center}
\begin{tabular}{lcc}
  \toprule
\multicolumn{1}{c}{\bf Model} &\multicolumn{1}{c}{\bf Embedding}  &\multicolumn{1}{c}{\bf Accuracy}\\
\multicolumn{1}{c}{\bf } &\multicolumn{1}{c}{\bf dimensions/topics}  &\multicolumn{1}{c}{\bf }
\\ \midrule
Paragraph vectors         &10000 & 93.0\% \\
LDA             &5000 & 82\% \\
Averaged word embeddings & 3000 & 84.9\% \\
Bag of words &  & 86.0\% \\
\bottomrule
\end{tabular}
\end{center}
\end{table}

\begin{figure}[h]
\centering
\includegraphics[width=.7\linewidth]{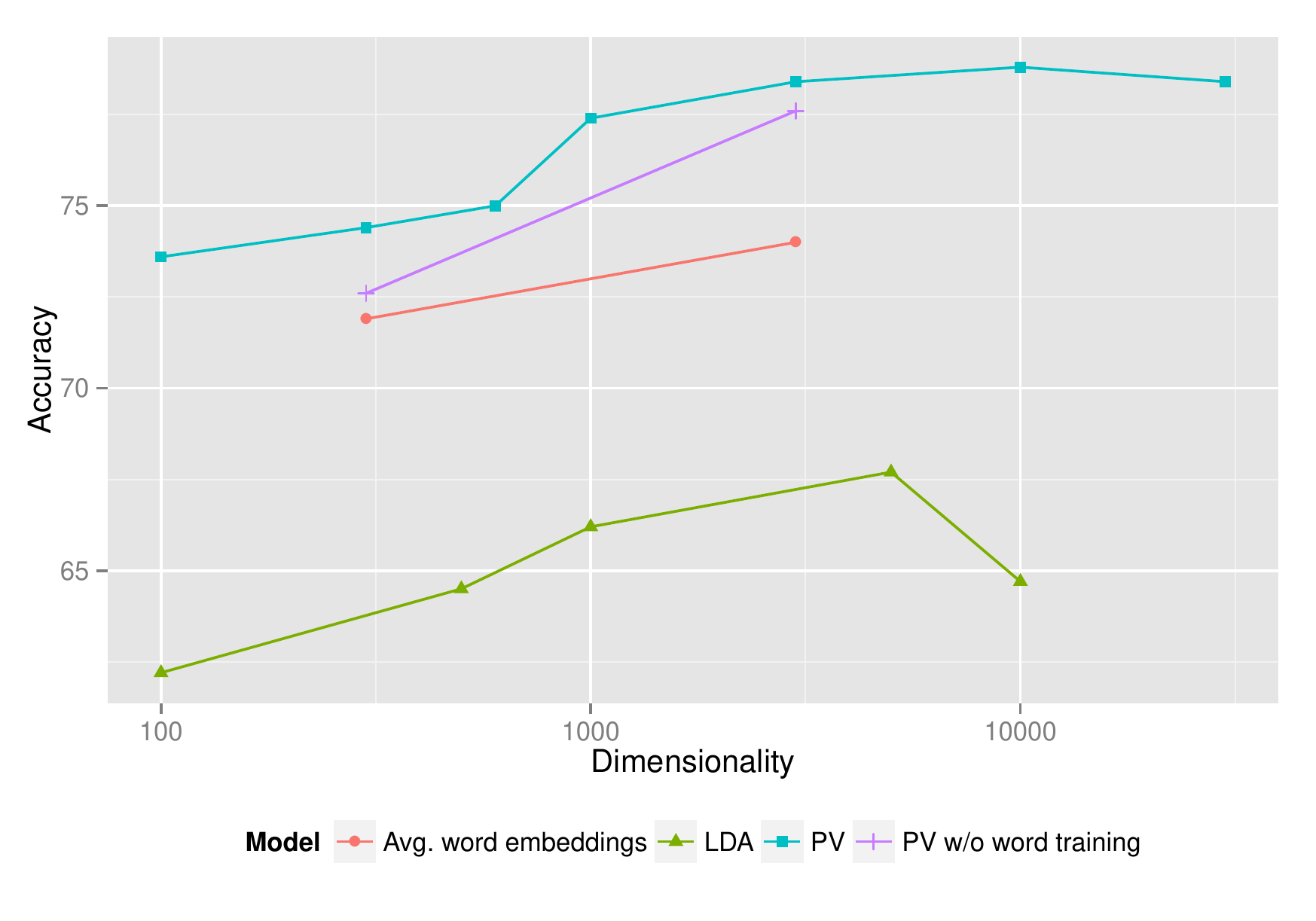}
\caption{Results of experiments on the generated Wikipedia triplet dataset. }
\label{wiki-generated-graph}
\end{figure}

\begin{table}[h!]
\caption{Performances of different methods on dataset with generated
  Wikipedia triplets on the best performing dimensionality.}
\label{wikipedia-generated-table}
\begin{center}
\begin{tabular}{lcc}
  \toprule
\multicolumn{1}{c}{\bf Model} &\multicolumn{1}{c}{\bf Embedding}  &\multicolumn{1}{c}{\bf Accuracy}\\
\multicolumn{1}{c}{\bf } &\multicolumn{1}{c}{\bf dimensions/topics}  &\multicolumn{1}{c}{\bf }
\\ \midrule
Paragraph vectors         &10000 & 78.8\% \\
LDA             &5000 & 67.7\% \\
Averaged word embeddings & 3000 & 74\% \\
Bag of words &  & 78.3\% \\
\bottomrule
\end{tabular}
\end{center}
\end{table}

From the results in Table~\ref{wikipedia-hand-table} and
\ref{wikipedia-generated-table}, it can be seen that paragraph vectors
perform better than LDA. We also see a peak in paragraph vector
performance at 10,000 dimensions. Both paragraph vectors and averaging word embeddings perform better than the
LDA model. For LDA, we found that TF-IDF weighting
of words and their inferred topic allocations did not affect the
performance. From these results, we can also see that joint training of word vectors improves the final quality of the paragraph vectors.

\newpage
\subsection{Performance of Paragraph Vectors on arXiv articles}
We extracted text from the PDF versions of 886,000 full arXiv
papers. In each case we only used the latest revision available. We
applied a minimum frequency cutoff to the vocabulary so that our final
vocabulary was 969,894 words. 

We performed experiments to find related articles using Paragraph
Vectors. In Table~\ref{arxiv-nn} and Table~\ref{arxiv-nn2}, we show
the nearest neighbours of the original Paragraph Vector paper
``Distributed Representations of Sentences and Documents'' and the
current paper. In Table~\ref{arxiv-vec}, we want to find the Bayesian
equivalence of the Paragraph Vector paper. This can be achieved by
vector operations: $pv$(``Distributed Representations of Sentences and
Documents'') - $wv$(``neural'') + $wv$(``Bayesian'') where $pv$ are
paragraph vectors and $wv$ are word vectors learned during the training
of paragraph vectors. The results suggest that Paragraph Vector works
well in these two tasks.

\begin{table}[h!]
\caption{arXiv nearest neighbours to ``Distributed Representations of Sentences and Documents'' using Paragraph Vectors.}
\label{arxiv-nn}
\begin{center}
\begin{tabular}{lc}
  \toprule
\multicolumn{1}{c}{\bf Title} &\multicolumn{1}{c}{\bf Cosine}\\
\multicolumn{1}{c}{} &\multicolumn{1}{c}{\bf Similarity}
\\ \midrule
        Evaluating Neural Word Representations in Tensor-Based Compositional Settings  & 0.771 \\ 
        Polyglot: Distributed Word Representations for Multilingual NLP & 0.764 \\
        Lexicon Infused Phrase Embeddings for Named Entity Resolution & 0.757 \\
        A Convolutional Neural Network for Modelling Sentences & 0.747 \\
        Distributed Representations of Words and Phrases and their Compositionality & 0.740 \\
        Convolutional Neural Networks for Sentence Classification & 0.735 \\
        SimLex-999: Evaluating Semantic Models With (Genuine) Similarity Estimation & 0.735 \\
        Exploiting Similarities among Languages for Machine Translation & 0.731 \\
        Efficient Estimation of Word Representations in Vector Space & 0.727 \\
        Multilingual Distributed Representations without Word Alignment & 0.721 \\
        \bottomrule
\end{tabular}
\end{center}
\end{table}

\begin{table}[h!]
\caption{arXiv nearest neighbours to the current paper using Paragraph Vectors.}
\label{arxiv-nn2}
\begin{center}
\begin{tabular}{lc}
  \toprule
\multicolumn{1}{c}{\bf Title} &\multicolumn{1}{c}{\bf Cosine}\\
\multicolumn{1}{c}{} &\multicolumn{1}{c}{\bf Similarity}
\\ \midrule
Distributed Representations of Sentences and Documents & 0.681 \\
Efficient Estimation of Word Representations in Vector Space & 0.680 \\
Thumbs up? Sentiment Classification using Machine Learning Techniques & 0.642 \\
Distributed Representations of Words and Phrases and their Compositionality & 0.624 \\
KNET: A General Framework for Learning Word Embedding using & 0.622 \\
~~~~~~~Morphological Knowledge & \\
Japanese-Spanish Thesaurus Construction Using English as a Pivot & 0.614 \\
Multilingual Distributed Representations without Word Alignment & 0.614 \\
Catching the Drift: Probabilistic Content Models, with Applications & 0.613 \\
~~~~~~~to Generation and Summarization & \\
Exploiting Similarities among Languages for Machine Translation & 0.612 \\
A Survey on Information Retrieval, Text Categorization, and Web Crawling & 0.610 \\
        \bottomrule
\end{tabular}
\end{center}
\end{table}

\begin{table}[h!]
\caption{arXiv nearest neighbours to ``Distributed Representations of
  Sentences and Documents'' - ``neural'' + ``Bayesian''. I.e., the
  Bayesian equivalence of the Paragraph Vector paper.}
\label{arxiv-vec}
\begin{center}
\begin{tabular}{lc}
  \toprule
\multicolumn{1}{c}{\bf Title} &\multicolumn{1}{c}{\bf Cosine} \\
\multicolumn{1}{c}{} &\multicolumn{1}{c}{\bf Similarity}
\\ \midrule
         Content Modeling Using Latent Permutations  & 0.629 \\
        SimLex-999: Evaluating Semantic Models With (Genuine) Similarity Estimation & 0.611 \\
        Probabilistic Topic and Syntax Modeling with Part-of-Speech LDA & 0.579 \\
        Evaluating Neural Word Representations in Tensor-Based Compositional Settings & 0.572 \\
        Syntactic Topic Models & 0.548 \\
        Training Restricted Boltzmann Machines on Word Observations & 0.548 \\
        Discrete Component Analysis & 0.547 \\
        Resolving Lexical Ambiguity in Tensor Regression Models of Meaning & 0.546 \\
        Measuring political sentiment on Twitter: factor-optimal design for & 0.544 \\
        ~~~~~~~multinomial inverse regression & \\
        Scalable Probabilistic Entity-Topic Modeling & 0.541 \\
        \bottomrule
\end{tabular}
\end{center}
\end{table}

To measure the performance of different models on this task, we
picked pairs of papers that had at least one
shared subject, the unrelated paper was chosen at random from papers
with no shared subjects with the first paper. We produced a dataset of 20,000 triplets by
this method.

\begin{figure}[h]
\centering
\includegraphics[width=.7\linewidth]{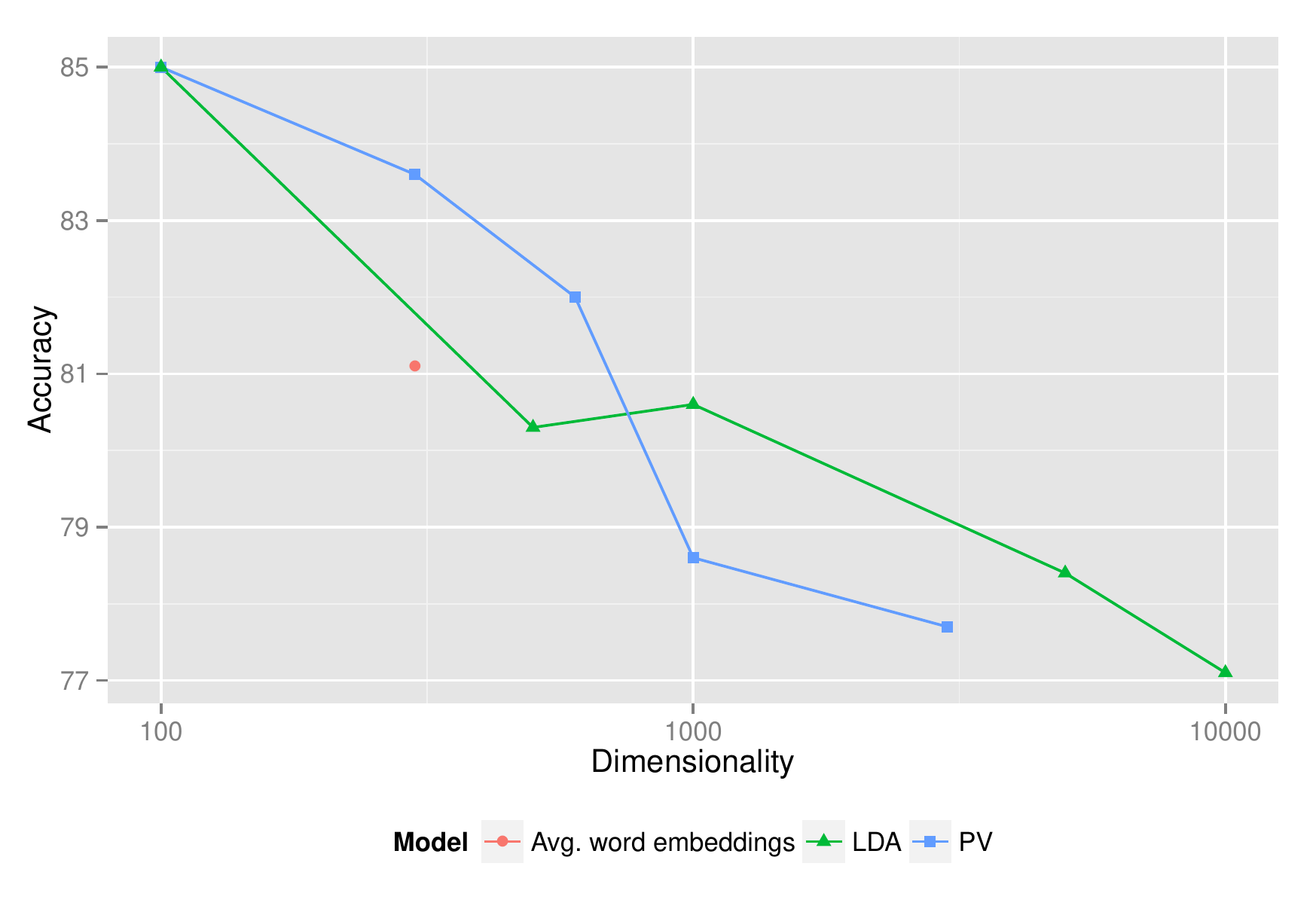}
\caption{Results of experiments on the arXiv triplet dataset. }
\label{wiki-generated-graph}
\end{figure}

\begin{table}[h!]
\caption{Performances of different methods at the best dimensionality on the arXiv
  article triplets.}
\label{arxiv-table}
\begin{center}
\begin{tabular}{lcc}
  \toprule
\multicolumn{1}{c}{\bf Model} &\multicolumn{1}{c}{\bf Embedding dimensions/topics}  &\multicolumn{1}{c}{\bf Accuracy}
\\ \midrule
Paragraph vectors         &100 & 85.0\% \\
LDA             &100 & 85.0\% \\
Averaged word embeddings & 300 & 81.1\% \\
Bag of words &  & 80.4\% \\
\bottomrule
\end{tabular}
\end{center}
\end{table}

From the results in Table~\ref{arxiv-table}, it can be seen that
paragraph vectors perform on par than the best performing number of topics for LDA. Paragraph Vectors are also less sensitive
to differences in embedding size than LDA is to the number of topics. We also see a peak in
paragraph vector performance at 100 dimensions. Both models perform
better than the vector space model. For LDA, we found that TF-IDF
weighting of words and their inferred topic allocations did not affect performance.

\section{Discussion}
We described a new set of results on Paragraph Vectors showing they
can effectively be used for measuring semantic similarity between long
pieces of texts. Our experiments show that Paragraph Vectors are
superior to LDA for measuring semantic similarity on Wikipedia articles across all sizes
of Paragraph Vectors. Paragraph Vectors also perform on par with LDA's best performing number of topics
on arXiv papers and perform consistently relative to the embedding size. Also surprisingly,
vector operations can be performed on them similarly to word
vectors. This can provide interesting new techniques for a wide range
of applications: local and nonlocal corpus navigation, dataset
exploration, book recommendation and reviewer allocation.

\bibliography{translate} 
\bibliographystyle{plain}
\end{document}